\documentclass{article} 
\usepackage{iclr2024_conference,times}

\usepackage{amsmath,amsfonts,bm}

\def\eqref#1{equation~\ref{#1}}

\def\1{\bm{1}}

\def\rvx{{\mathbf{x}}}

\def\rvz{{\mathbf{z}}}

\DeclareMathAlphabet{\mathsfit}{\encodingdefault}{\sfdefault}{m}{sl}
\SetMathAlphabet{\mathsfit}{bold}{\encodingdefault}{\sfdefault}{bx}{n}

\usepackage{hyperref}
\usepackage{url}
\usepackage{graphicx}
\usepackage{svg}
\usepackage{amsmath}
\usepackage{amssymb}
\usepackage{booktabs}
\usepackage{wrapfig}
\usepackage[accsupp]{axessibility} 

\title{Towards Climate Variable Prediction\\ with Conditioned Spatio-Temporal Normalizing Flows}

\author{Christina Winkler  \\
Mila - Quebec AI Institute\\
\texttt{christina.winkler@mila.quebec} \\
\And
David Rolnick \\
McGill University and Mila - Quebec AI Institute\\
\texttt{drolnick@mila.quebec} \\
}

\iclrfinalcopy 
\begin{document}

\maketitle

\begin{abstract}
This study investigates how conditional normalizing flows can be applied to remote sensing data products in climate science for spatio-temporal prediction. The method is chosen due to its desired properties such as exact likelihood computation, predictive uncertainty estimation and efficient inference and sampling which facilitates faster exploration of climate scenarios. Experimental findings reveal that the conditioned spatio-temporal flow surpasses both deterministic and stochastic baselines in prolonged rollout scenarios. It exhibits stable extrapolation beyond the training time horizon for extended rollout durations. These findings contribute valuable insights to the field of spatio-temporal modeling, with potential applications spanning diverse scientific disciplines.
\end{abstract}

\section{Introduction}
Global Climate Models (GCMs) describe physical processes by partial differential equations. These are often computationally expensive due to coarse grid discretization and solving over large spatial and temporal domains. Yet, there's a growing need for precise forecasts of Earth's climate, not just globally but also locally. This is crucial for guiding local responses to extreme weather events and implement adaptive strategies for e.g. water management or agricultural practices. To bridge this disparity, the machine learning community contributed vastly with deep learning based methods for climate variable super resolution to obtain high-resolution simulation data \citep{Wang_2018_ECCV_Workshops, watson2020investigating,Groenke2020-il, Singh2019DownscalingNW, harder2022generating, watson2020investigating, Chaudhuri2020}. Moreover, researchers are drawing inspiration from video prediction techniques utilizing deep learning, including recurrent neural networks (RNNs) and convolutional neural networks (CNNs), as well as stochastic approaches to accelerate the emulation of future states of remote sensing variables \citep{Rasp2018NeuralNF, Reichstein2019DeepLA, Rasp2018, Ravurietal2021, Klemmer2021, Saxena2019DGANDG, Zhu2020InterpretableDG, BenBouallegue2023ImprovingME, Ali2021ExploitingDS, Amato2020ANF, wang2022, FERCHICHI2022101552, Pisl2023, Drees2022TimeDI, Ruwurm2018ConvolutionalLF, Gao2020GenerativeAN, Ji2023, Ravuri2021}. This approach can facilitate faster exploration of climate scenarios and potentially enhance the efficiency of GCM-based climate research and decision-making processes. These contributions could allow researchers to incorporate subgrid convective processes in climate models, allowing for more accurate predictions of thunderstorms, heavy rainfall events and tropical cyclones.

In this work, we propose conditional normalizing flows for spatio-temporal modelling of remote sensing data products for climate science. Specifically, we investigate the capabilities of conditional normalizing flows for climate variable forecasting of different temporal and spatial resolutions. Invertible stochastic models allow for exact likelihood computation, predictive uncertainty estimation and fast inference and sampling. Flows for video modelling have yet been explored \citep{Kumar2019VideoFlow, zand2022flowbased, Davtyan2022}. Different from the literature, in our approach temporal correlations are learned via conditioning on a compressed memory state produced by a convolutional gated LSTM \citep{DauphinFAG16} of the input frame.  Our main contribution can be summarized as follows:

\begin{itemize}
    \item We introduce a novel method for efficient spatio-temporal modelling with conditioned normalizing flows. 
    \item We employ ST-Flow on ERA5 data, revealing its consistent and stable performance over long rollout durations, as measured by the RMSE metric, when compared against both deterministic and stochastic baselines. 
\end{itemize}

\section{Method}\label{sec:method}

A Normalizing Flow in continuous space is based on the change of variables formula. Given two spaces of equal dimension $\mathcal{Z}$ and $\mathcal{X}$; a diffeomorphism $f_\phi : \mathcal{X} \rightarrow \mathcal{Z}$ and a prior distribution $p_z(\mathbf{z})$ which is easy to sample from, we can model a complicated distribution $p_X(\mathbf{x})$ as

\begin{equation}
    p_X(\rvx) = p_Z(f_{\phi}(\rvx)) \left | \frac{\partial f_\phi(\rvx)}{\partial \rvx} \right |. \label{eq:change-of-variables}
\end{equation}

Here $\left | \partial f_\phi(\rvx) / \partial \rvx \right |$ is the determinant of the Jacobian of the transformation between $\rvz$ and $\rvx$, evaluated at $\rvx$, accounting for volume changes in $p_z(\rvz)$ induced by $f_\phi$. Typically, the transformation $f_\phi$ is defined as composition of diffeomorphisms, i.e.  
$f=f_0 \circ f_1 \circ ... \circ f_K$. 

\subsection{Conditioned Spatio-Temporal Normalizing Flow}\label{sec:st-cnf}
We propose using conditional normalizing flows \citep{winkler2019, Lugmayr2021NormalizingFA} for spatio-temporal sequence prediction. Invertible stochastic methods are particularly desirable for the task of climate variable prediction due to their ability to capture predictive uncertainty, ability to generate synthetic data for scenario analysis and handling missing data and imputation. Assume we have a target frame $\rvx_t$ and a context frame which can be arbitrarily long $(\rvx_{t-1}, \dots , \rvx_{T})$ at the $t$-th time-step such that we have the training tuple $(\rvx_t; \rvx_{<t}) \in \{X_t\}$. We learn a complicated distribution $ p_\rvx(\rvx_t \vert \rvx_{<t})$ using a conditional prior $p_\theta (\rvz_t \vert \rvz_{<t})$ and a mapping $f_\phi: \mathcal{X}^t \times \mathcal{X}^{<t} \rightarrow \mathcal{Z}^t$, which is bijective in $\mathcal{X}^t$ and $\mathcal{Z}^t$. The likelihood of this model can then be written as:
\begin{equation}
    \begin{aligned}
        p_{\rvx} (\rvx_t \vert \rvx_{<t}) &= p_{\theta} (\rvz_{t} \vert \rvz_{<t}) \left | \det \frac{\partial \rvz_t}{\partial \rvx_{t}} \right | 
        = p_\theta ( f_\phi(\rvx_t, \rvx_{<t}) \vert \ \rvz_{<t}) \left | \det \frac{\partial f_\phi(\rvx_{t}, \rvx_{<t})}{\partial \rvx_t} \right | .\\
    \end{aligned}
\end{equation}

\begin{wrapfigure}{r}{0.5\textwidth}
\includesvg[width=0.5\columnwidth]{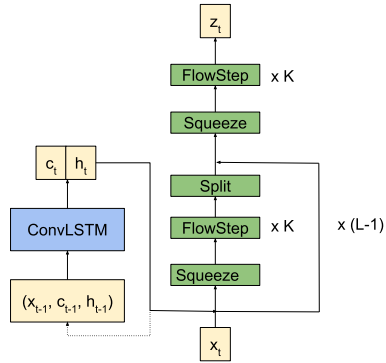}
\caption{\textit{ST-Flow architecture with K flow steps and L scales. For further explanation, see Appendix A.}}\label{fig:ms-arch}
\end{wrapfigure}

Over the temporal axis, the latent prior can then be factorized as:

\begin{equation}
    p_\theta(\rvz | \mathbf{h}) = \prod_{t=1}^T p_\theta (\rvz_t \vert \mathbf{h}_{t}),
\end{equation}

where $\mathbf{h}= \textit{GatedConvLSTM}(x_{t-1},x_{t-2}, ... )$ represents the hidden state at time $t$. We chose a gated convolutional LSTM due to its ability to efficiently capture long-range dependencies in sequential data while addressing the vanishing gradient problem, which is crucial for tasks requiring robust representation of past states and inputs. Factorization of hierarchical latent variables $\{\rvz_t^l\}_{l=1}^{L}=f_\phi(\rvx_t, \rvx_{<t})$ over the scales can then be written as:

\begin{equation}
    p_\theta(\rvz_t \vert \mathbf{h}_{t}) = \prod_{l=1}^L p_\theta (\rvz_t^{(l)} \vert \rvz_t^{(>l)}, \mathbf{h}_{t}), \\
\end{equation}

where $\rvz_{t}^{(>l)}$ is the set of latent variables at the same time step at higher scales. At each scale \textit{l}, the split prior $p_\theta (\rvz_t^{(l)} \vert \rvz_t^{(>l)}, \mathbf{h}_{t})$ takes on the form of a conditionally factorized Gaussian density: 

\begin{equation}
    \begin{aligned}
           p_\theta (\rvz_t^{(l)} \vert \rvz_t^{(\geq l)}, \mathbf{h}_{t}) &= \mathcal{N}(\rvz_t^{(l)} ; \mu, \sigma)\\
       \text{where } \mu, \log (\sigma) &= \text{NN}_\theta(\rvz_t^{(\geq l)}, \mathbf{h}_t),\\
    \end{aligned}
\end{equation}

where $\text{NN}_\theta$ is a CNN to predicting $\mu$ and $\log \sigma$. The learning objective of the full model with memory-based latent dynamics can then be formulated as:

\begin{equation}
    \begin{aligned}
        \log p_\rvx(\rvx \vert \mathbf{h}) = \log p_\theta (\rvz \vert \mathbf{h}) + \sum_{k=1}^K \log \left | \det \frac{\partial f_k(\rvz_k; \mathbf{h})}{\partial f_{k-1}(\rvz_{k-1}, \mathbf{h})} \right |,\\
    \end{aligned}
\end{equation}

where \textit{K} represents the number of flow steps. An overview of the model architecture is provided in Figure \ref{fig:ms-arch}.

\section{Experiments}\label{sec:experiments}

\begin{figure}[tb]
\centering
\minipage{0.4\columnwidth}
   \includegraphics[width=0.8\columnwidth]{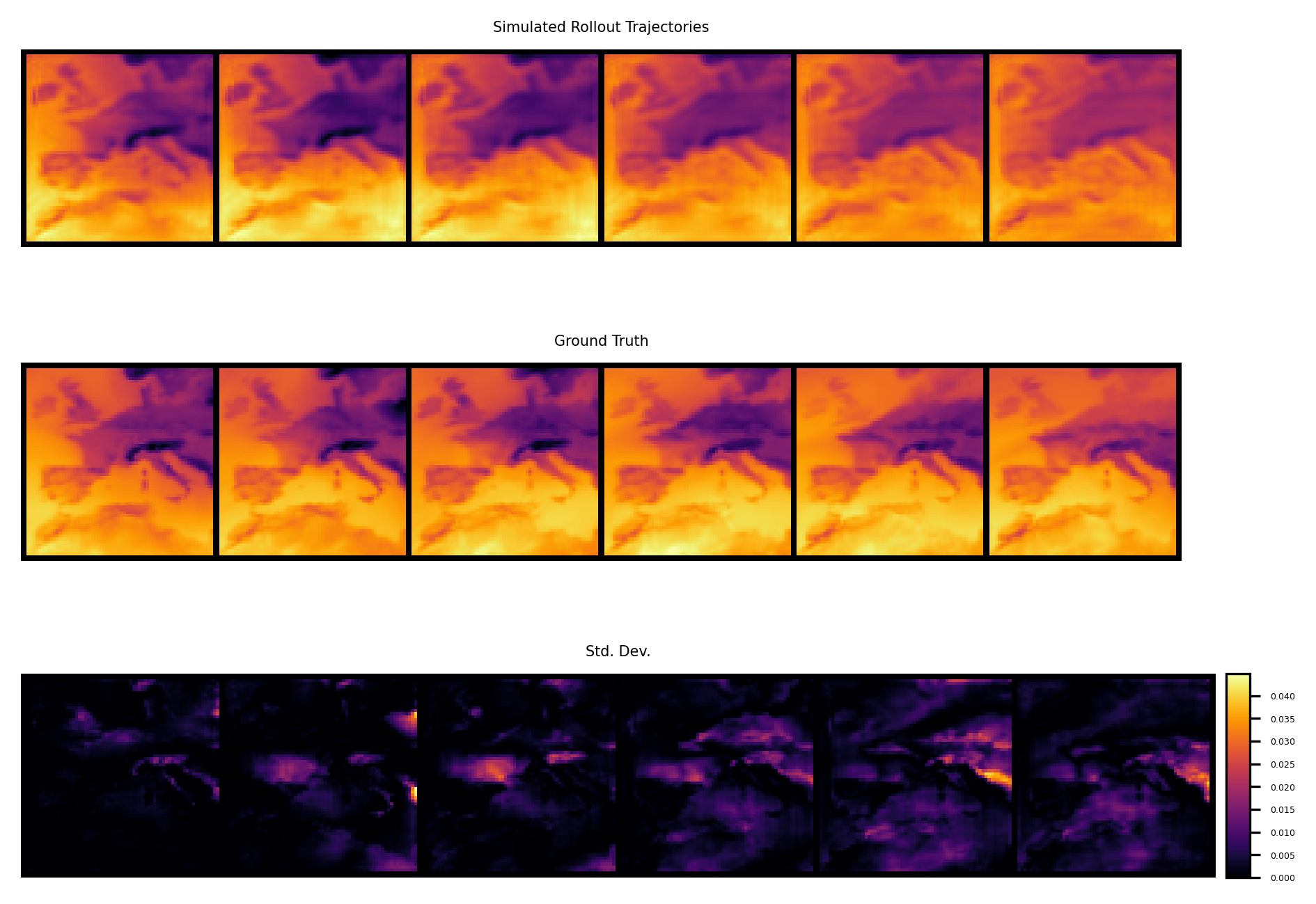}
  \text{$\qquad$ $\qquad$ \tiny \textit{Temperature}}
 \endminipage 
 \minipage{0.4\columnwidth}
   \includegraphics[width=\columnwidth]{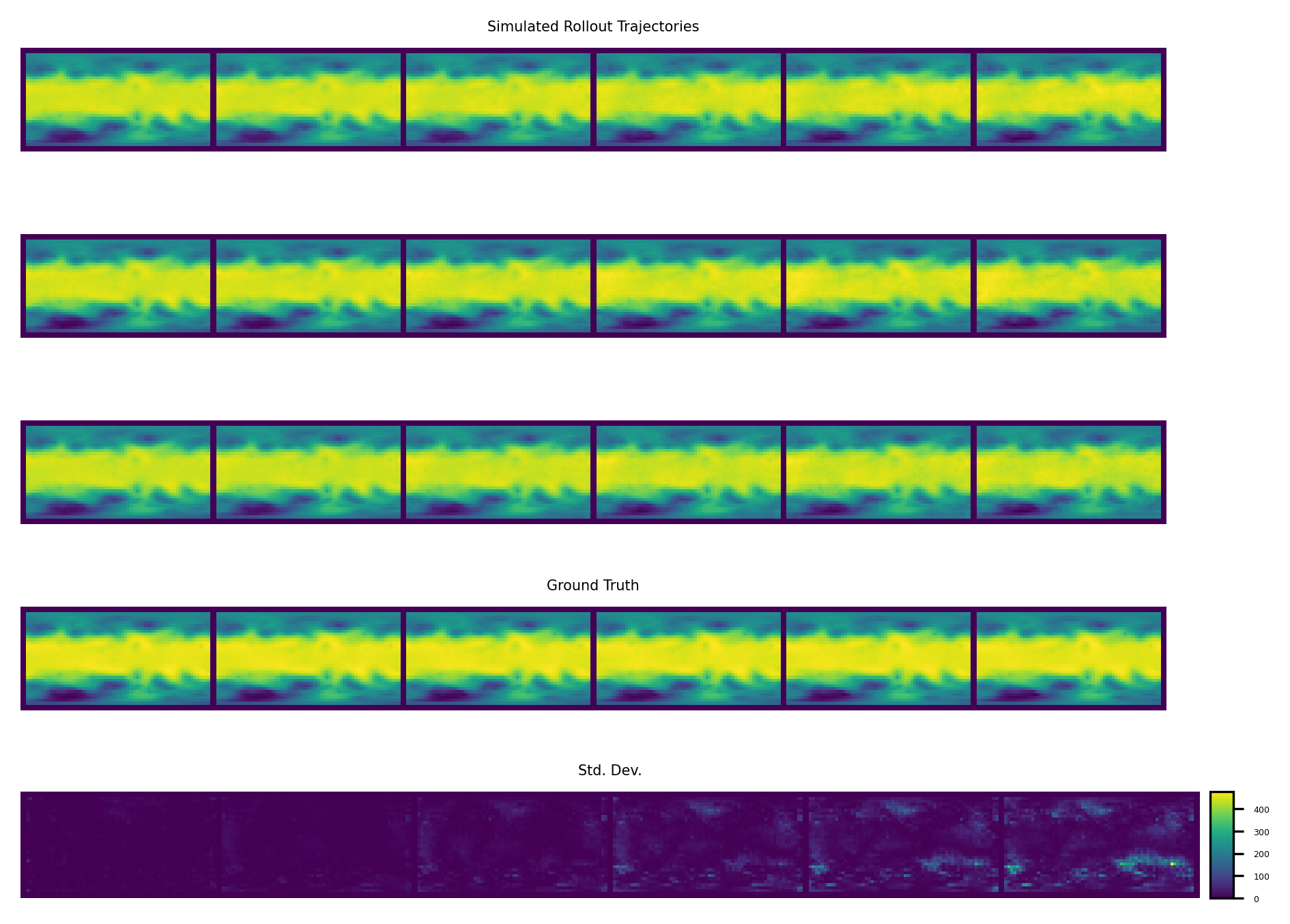}
   \text{$\qquad$$\qquad$ \tiny \qquad \quad \textit{Geopotential}}
 \endminipage 
 \caption{\textit{Visualization of rollout trajectories starting from the same initial conditions on the ERA5 temperature and geopotential dataset from the conditional normalizing flow model. The last row shows the squared absolute error. For rollouts from other methods, see appendix.}}
 \label{fig:rollouts}
 \end{figure}
 
We conduct experiments on ERA5 remote sensing data products, which are produced by the European Centre for Medium-Range Weather Forecasts (ECMWF). ERA5 data is generated using state-of-the-art numerical weather prediction models, assimilating a wide range of observational data sources, such as satellites, weather stations, and ocean buoys. Figure \ref{tab:data-summary} describes a summary of the datasets used in this study. For preprocessing, we transform values $Z$ by $X=\frac{Z-\min{Z}}{\max{Z}-\min{Z}}$ such that they lie within range [0,1]. We employ a learning rate of 2e-4 using a Step-Wise learning rate scheduler with a decay rate of 0.5 after every 200000th parameter update step. We used the Adam optimizer \citep{kingma2014method} with exponential moving average and coefficients of running averages of gradients and its square are set to $\beta=(0.9,0.99)$. 

\begin{figure}[htb]
\begin{center}
\resizebox{0.6\linewidth}{!}{
\begin{tabular}{|c|c|c|c|}
\hline
\textbf{Dataset} & \textbf{Resolution} & \textbf{Time Span} & \textbf{Training/Validation/Test Samples} \\
\hline
T2M & 0.25° & 1979 - 2020 & 1193 / 340 / 170 \\
\hline
500 hPa & 5.625° & 1979 - 2017 & 298,043 / 17,532 / 35,064 \\
\hline
\end{tabular}}
\end{center}
\caption{\textit{Dataset summary with their resolutions, time spans, and training/validation/test samples.}}\label{tab:data-summary}
\end{figure}



\subsection{Quantitative Results}
\begin{figure}[t]
\centering
\minipage{0.45\columnwidth}
   \includegraphics[width=1.1\columnwidth]{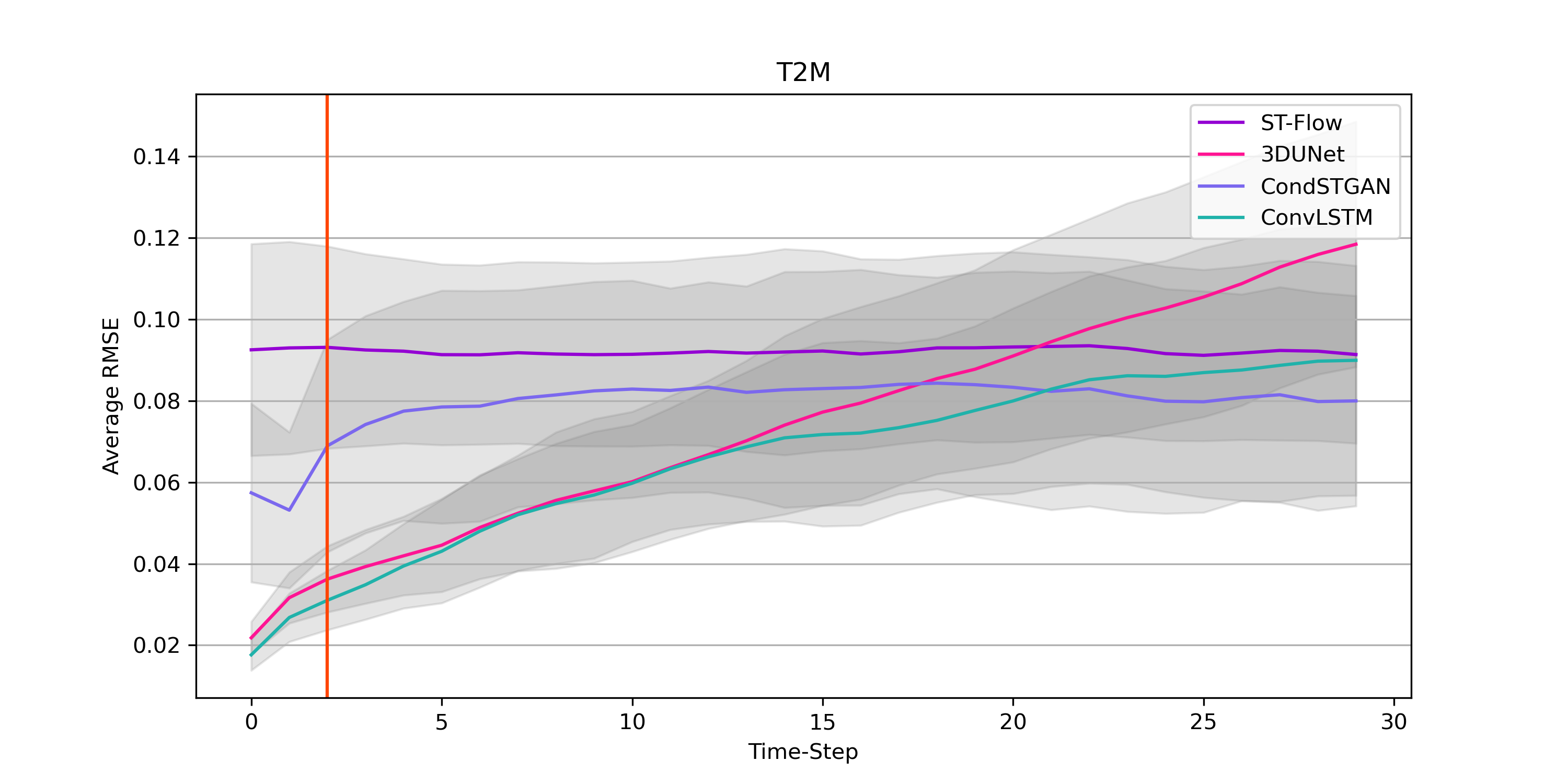}
   \centering
   (a) \textit{T2M}
 \endminipage 
 \minipage{0.45\columnwidth}
   \includegraphics[width=1.1\columnwidth]{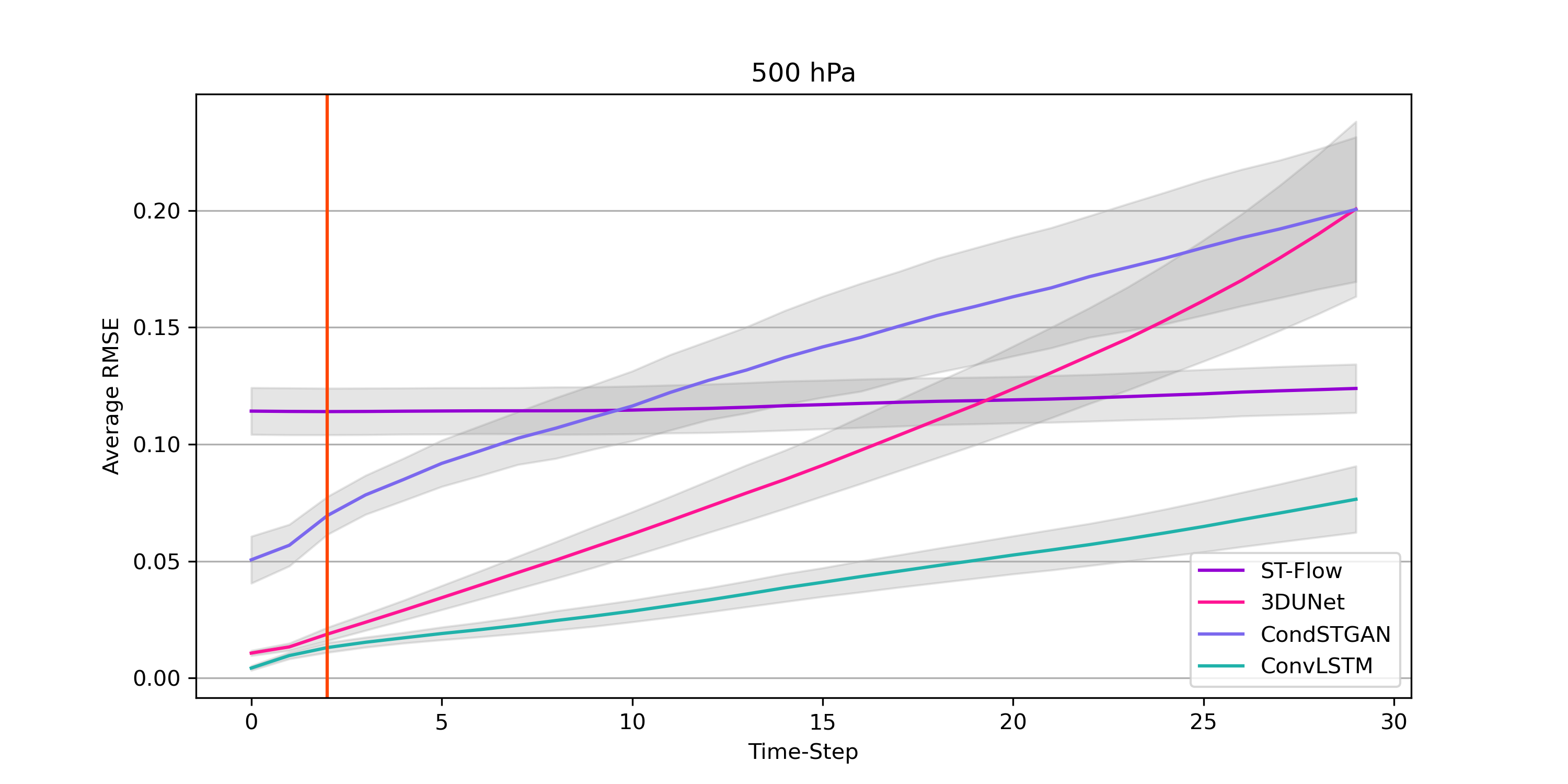}
   \centering
   (b) \textit{500 hPa}
 \endminipage 
 \caption{\textit{RMSE curves over 100 test samples on the ERA5 daily temperature and hourly geopotential dataset computed for different models. The vertical line indicates the length of the context window size during training, which we set to 2.}}
 \label{fig:rmse-curves-model-comp}
 \end{figure}

In Figure \ref{fig:rmse-curves-model-comp} we present a comparative analysis of the ST-Flow against two deterministic approaches (3DUNet and ConvLSTM), as well as a spatio-temporal conditional GAN serving as stochastic baseline using the RMSE metric. All models were trained with the same parameter budget and on the original input resolutions of the respective datasets for a lead time of 30 time-steps. Examining the results on the hourly geopotential dataset (Fig. \ref{fig:rmse-curves-model-comp} a)) reveals an initial advantage for the deterministic methods over the stochastic ones in terms of performance. However, as the rollout trajectory progresses, the ST-Flow model consistently maintains the most stable RMSE scores throughout the entire duration. Conversely, the conditional GAN architecture exhibits the poorest performance among the models evaluated. On the daily temperature dataset (Fig. \ref{fig:rmse-curves-model-comp} b)), results indicate that the deterministic methods perform better initially, but decrease in performance for longer rollout durations. The stochastic methods provide more stable results over the whole rollout period. On the hourly dataset, the variability in performance is lower than for the coarser temporally resolved temperature dataset for all models.

The initial advantage for deterministic methods over stochastic ones, such as observed in the hourly geopotential dataset, can be attributed to their ability to effectively capture short-term patterns and dependencies within the data. Deterministic methods excel at directly predicting the next value in a sequence based on the current and past observations, which can be advantageous in scenarios where short-term trends dominate the data. However, as the rollout trajectory progresses, deterministic methods may struggle to maintain accurate predictions over longer horizons due to their inherent limitations in capturing complex, evolving patterns and dependencies. These methods may become increasingly susceptible to errors accumulating over time, leading to a degradation in performance.

The conditional GAN provides superior performance over the ST-Flow on the temperature dataset, but not on the geopotential dataset. This may be attributed to the simpler temporal and spatial dynamics in the temperature dataset as compared to the geopotential dataset.

In contrast, the ST-Flow model, demonstrates greater stability in RMSE scores over the entire duration of the rollout trajectory on both datasets. By leveraging its capacity to capture spatio-temporal long-term relationships, the ST-Flow model can make more robust and consistent predictions over extended horizons compared to deterministic methods.

Overall, the observed trends underscore the importance of considering the modeling capabilities of different approaches, as well as the specific characteristics and dynamics of the dataset, when assessing performance over varying prediction horizons.

 

\subsection{Efficient Spatial Representation Learning for Climate Variable Forecasting with Conditional Normalizing Flows}
\begin{figure}[]
\centering
\minipage{0.45\columnwidth}
   \includegraphics[width=\columnwidth]{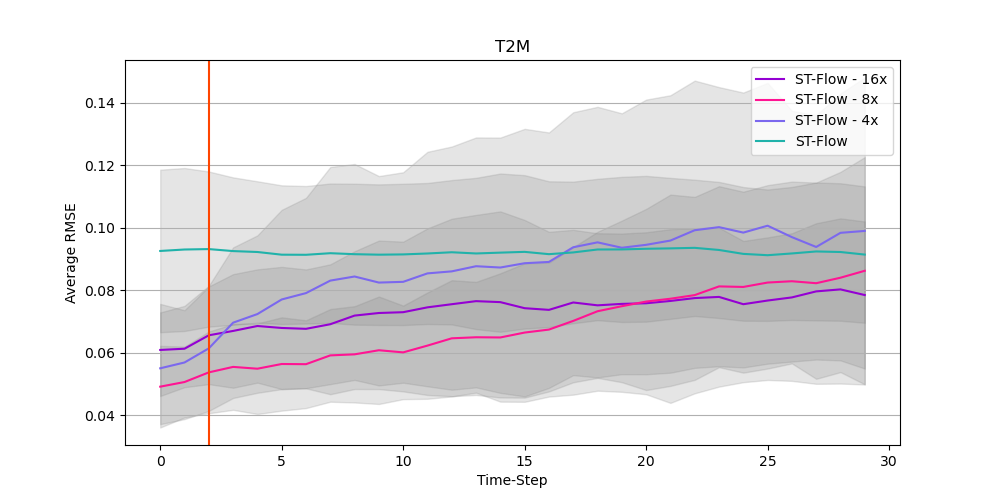}
   \centering
   (a) \textit{T2M}
 \endminipage 
 \minipage{0.45\columnwidth}
   \includegraphics[width=\columnwidth]{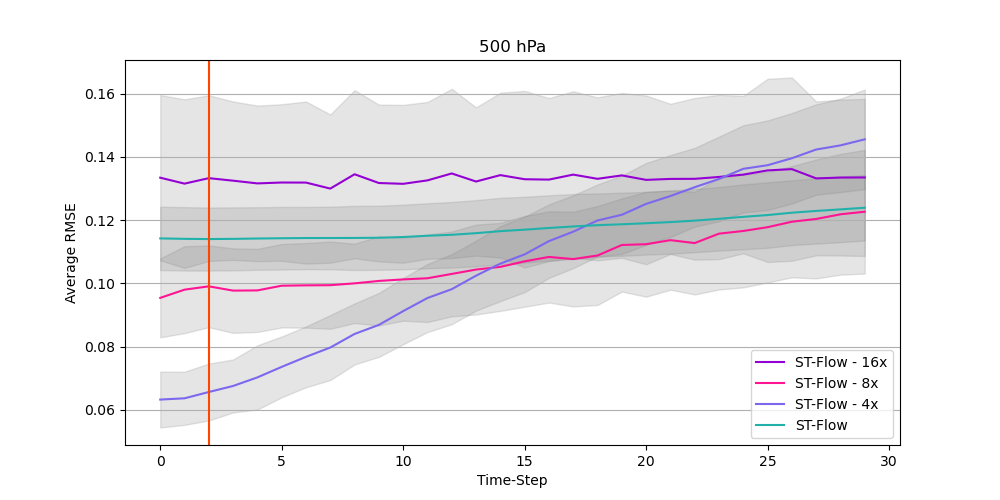}
   \centering
   (b) \textit{500 hPa}
 \endminipage 
 \caption{\textit{RMSE curves over 100 test samples on the ERA5 daily temperature and hourly geopotential dataset for different input resolutions of the data (orig,4x,8x,16x). The vertical line indicates the length of the context window size during training, which we set to 2.}}
 \label{fig:rmse-curves-stflow-compressed}
 \end{figure}

In this experiment, we leverage the power of conditional normalizing flows as auto-encoders to efficiently reduce the spatial dimensionality of input frames. Subsequently, we employ simulations based on these condensed representations, providing a streamlined method for handling large-scale climate datasets. The outcomes, depicted in Figure \ref{fig:rmse-curves-stflow-compressed}, present the RMSE results of the spatio-temporal normalizing flow across a spectrum of downsampling resolutions (4x, 8x, 16x) applied during the simulation. These RMSE values are computed within the reconstructed original image space, achieved through the training of a conditional normalizing flow as post-processing operation.

Generally, findings from both datasets indicate that for longer rollout durations, employing a coarser spatial representation yields more stable outcomes. The phenomenon of employing a coarser spatial representation leading to more stable outcomes for longer rollout durations can be attributed to several factors. Firstly, by simplifying the system's dynamics through coarsening, the complexity of the simulation is reduced, making it more resilient to small-scale fluctuations or uncertainties that may arise over time. Secondly, the smoothing effect of coarsening, achieved through averaging or aggregating spatial information, helps to dampen out noise or variability at smaller scales, resulting in a smoother and more predictable system behavior. Additionally, the computational efficiency gained from using a coarser spatial representation enables longer simulations to be conducted more efficiently, allowing researchers to study the system's behavior over extended time periods without significant computational overhead. 

However, performance in relation to the downsampling factor varies across datasets, likely due to discrepancies in spatial and temporal resolutions. For instance, datasets with higher spatial resolutions such as the geopotential dataset may contain more detailed information about the system's dynamics, making it challenging to effectively condense the data without losing critical features. This is also reflected in the results presented in Figure \ref{fig:rmse-curves-stflow-compressed} panel b).

The efficacy of dimensionality reduction on input frames lies in its ability to eliminate redundant information while retaining essential data necessary for subsequent prediction tasks. By reducing the dimensionality of the dataset, the model focuses on capturing the most relevant features, leading to improved efficiency and interpretability. This process helps in discarding extraneous details that may introduce noise or complexity without adding significant value to the predictive task at hand. As a result, the dimensionality reduction enhances the model's ability to generalize patterns and make accurate predictions, even when operating on lower-dimensional representations of the input data.





\section{Conclusion} 
Experimentally, we have shown that deterministic methods initially outperform stochastic ones due to their proficiency in capturing short-term patterns, notably in the geopotential dataset. However, their performance declines over longer horizons due to limitations in capturing evolving patterns. Overall, experiments with the ST-Flow show that simplifying the spatial representation of the system aids in stabilizing its behavior over longer durations by reducing complexity, smoothing variability and improving computational efficiency. In summary, while the existing ERA5 datasets may be a great proof of concept, it is important to evaluate the trade-offs between computational complexity, data accuracy, and the specific needs of the real-world modeling task. Also, there is a need for exploring model calibration techniques such as incorporating physics-informed neural networks to ensure that predictions of AI models accurately reflect the true underlying data distribution and account for any systematic biases or errors.

\section{Acknowledgments}
This research was supported by compute resources, software, and technical help generously provided by the Mila - Quebec AI Institute (\url{mila.quebec}). 


\clearpage
{\small
\bibliographystyle{iclr2024_conference}
\bibliography{bibliography}
}

\newpage
\appendix
\section{Appendix: Simulated Rollout Trajectories} \label{appdx:rollouts}

\begin{figure}[]
\centering
\minipage{0.4\columnwidth}
   \includegraphics[width=0.8\columnwidth]{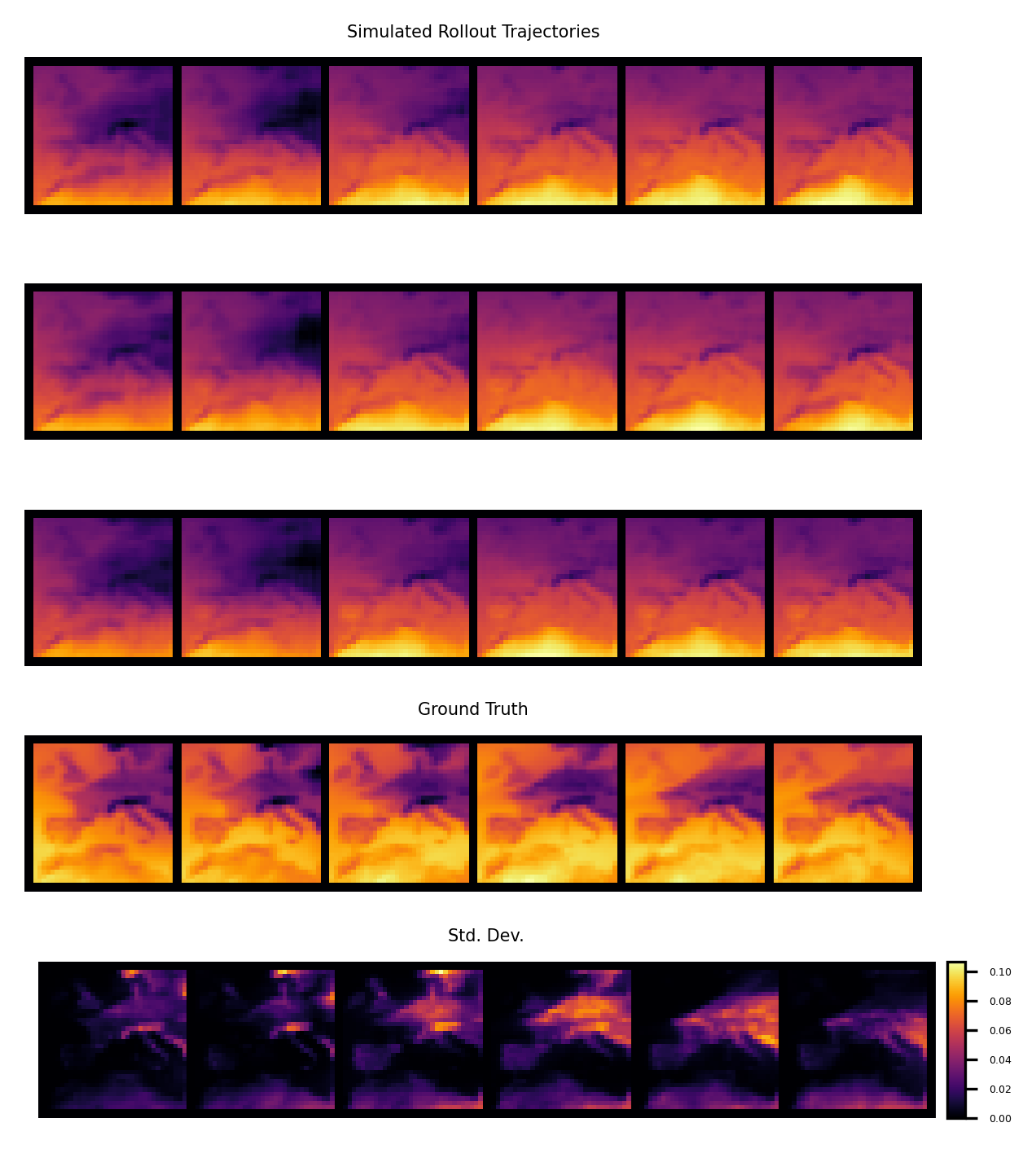}
  \text{$\qquad$ $\qquad$\tiny \textit{Temperature}}
 \endminipage 
 \minipage{0.4\columnwidth}
   \includegraphics[width=\columnwidth]{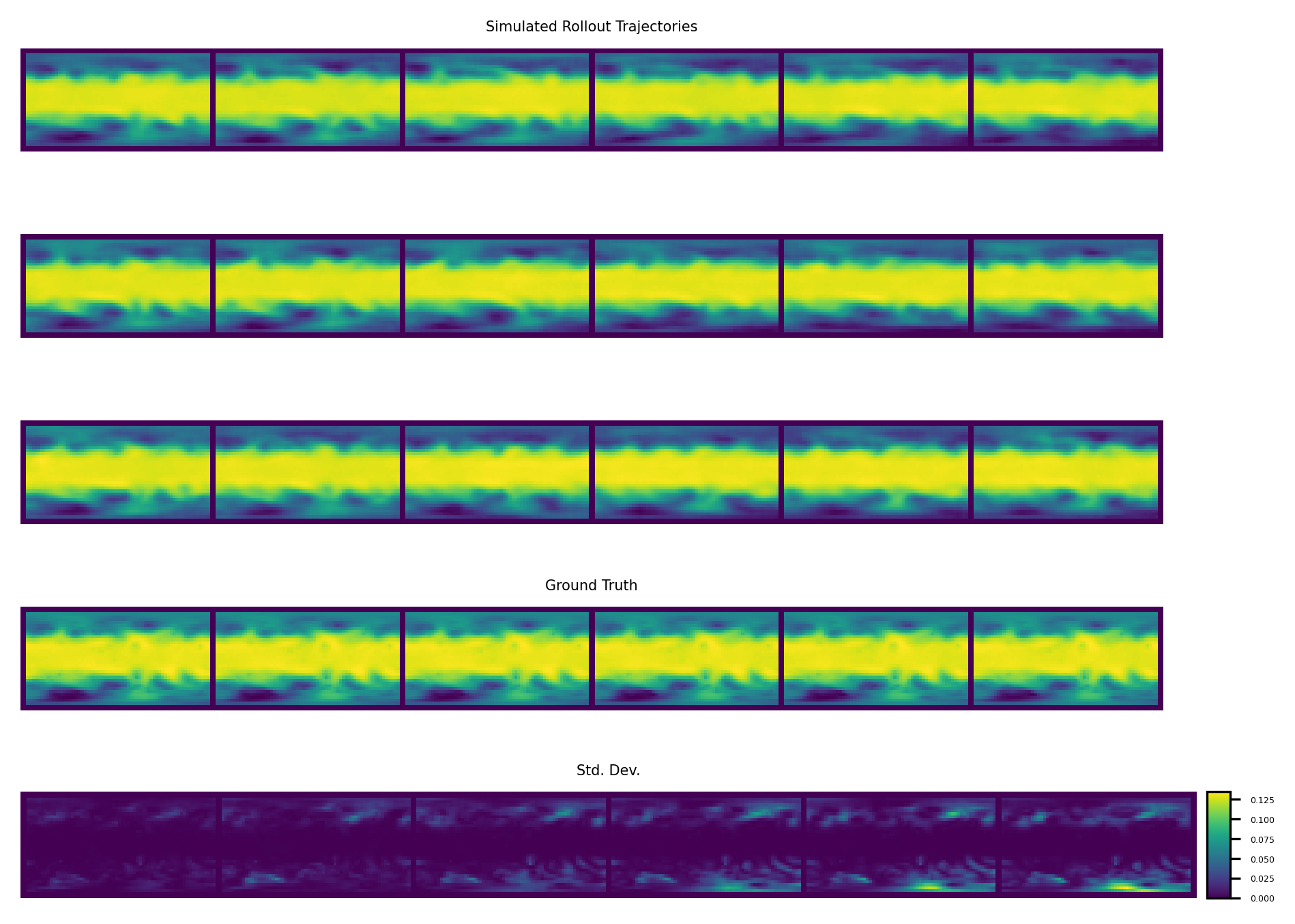}
   \text{$\qquad$$\qquad$ \tiny \textit{Geopotential}}
 \endminipage 
 \caption{\textit{Visualization of rollout trajectories starting from the same initial conditions on the ERA5 temperature and geopotential dataset from the \textbf{conditional GAN} model. The last row shows the squared absolute error. For rollouts from other methods, see appendix.}}
 \label{fig:rolloutsgan}
 \end{figure}

\begin{figure}[]
\centering
\minipage{0.4\columnwidth}
   \includegraphics[width=0.8\columnwidth]{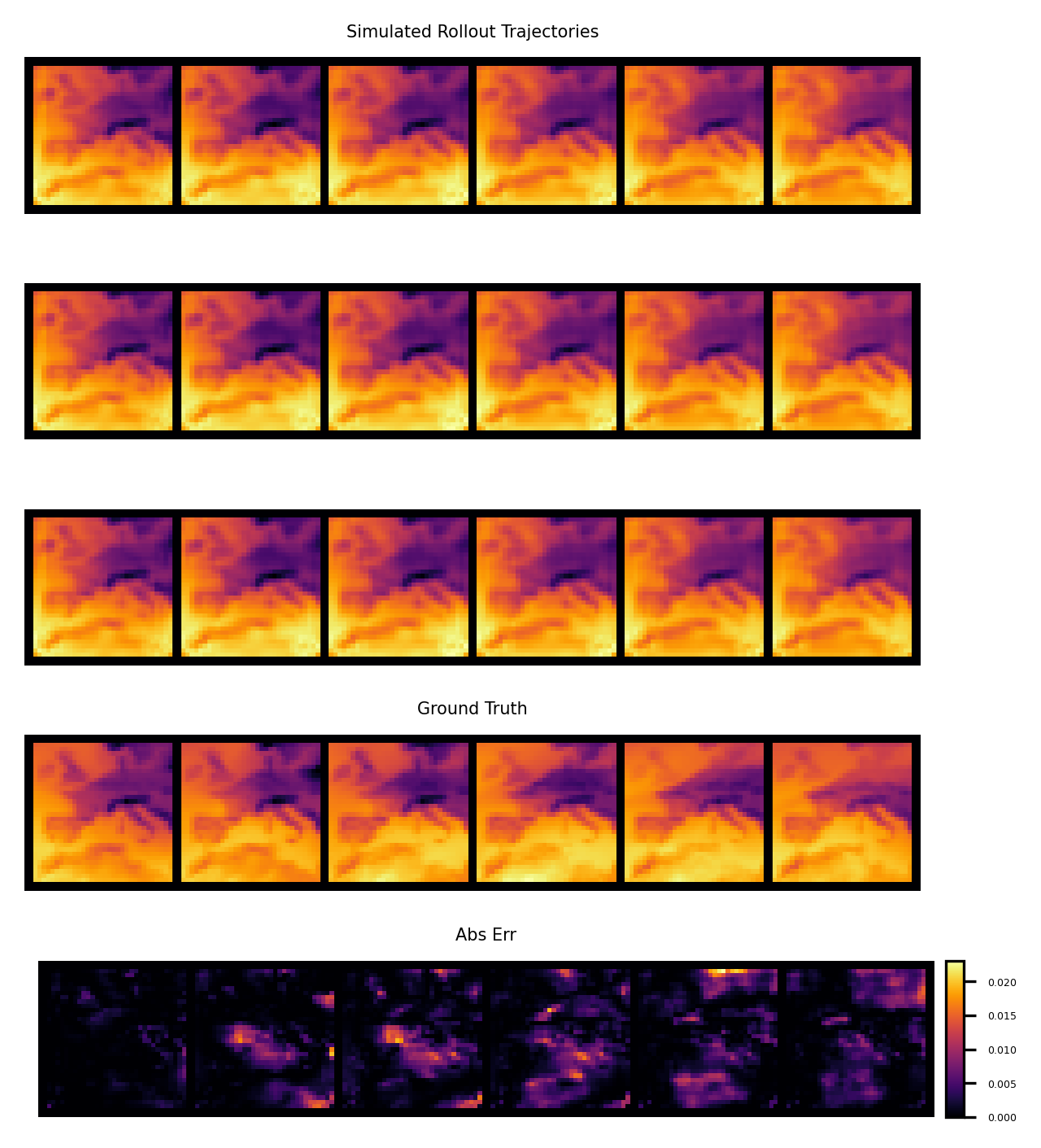}
  \text{$\qquad$$\qquad$ \tiny \textit{Temperature}}
 \endminipage 
 \minipage{0.4\columnwidth}
   \includegraphics[width=\columnwidth]{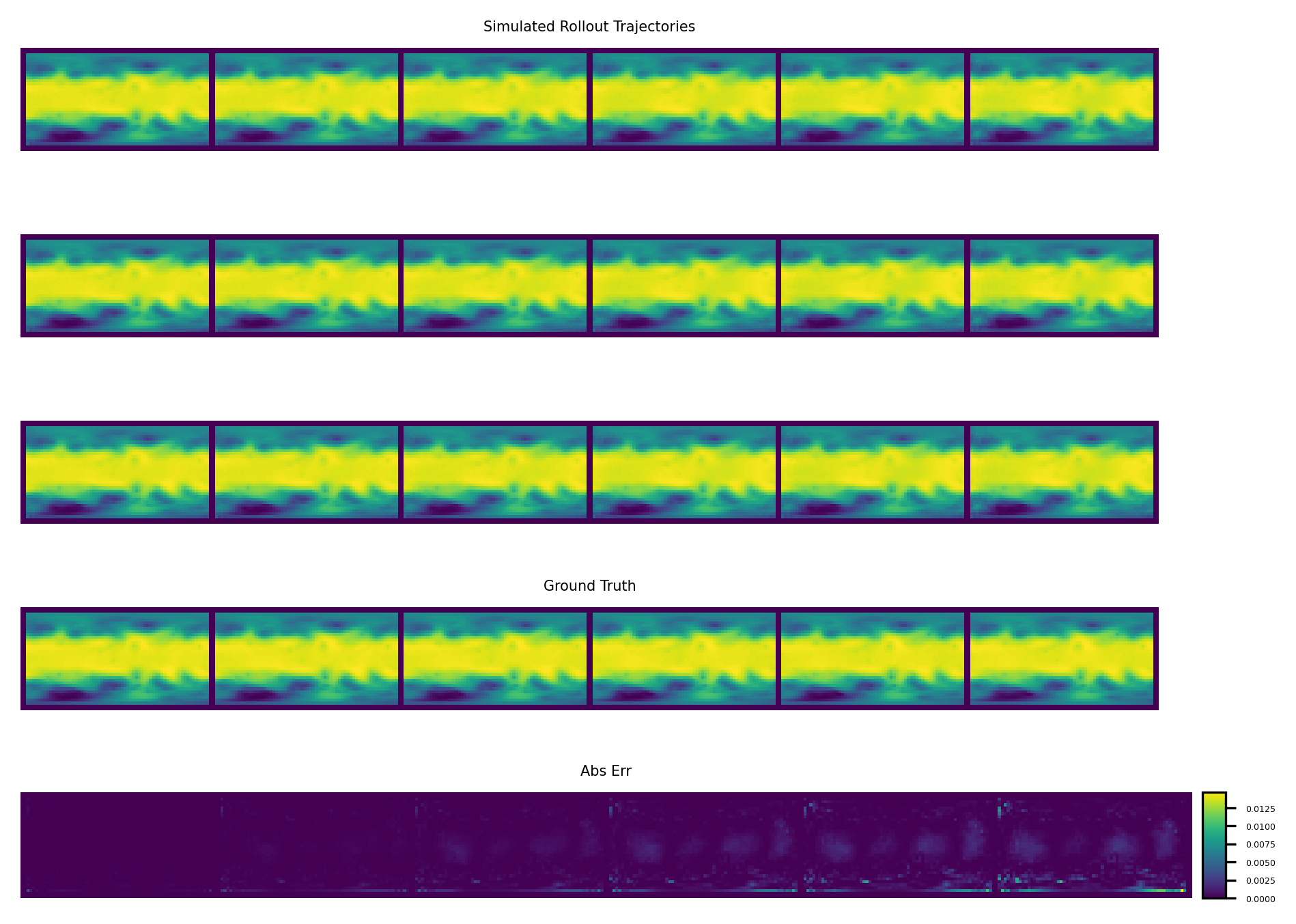}
   \text{$\qquad$ $\qquad$\tiny \textit{Geopotential}}
 \endminipage 
 \caption{\textit{Visualization of rollout trajectories starting from the same initial conditions on the ERA5 temperature and geopotential dataset from the \textbf{convLSTM} model. The last row shows the squared absolute error. For rollouts from other methods, see appendix.}}
 \label{fig:rolloutsconvlstm}
 \end{figure}

\begin{figure}[]
\centering
\minipage{0.4\columnwidth}
   \includegraphics[width=0.8\columnwidth]{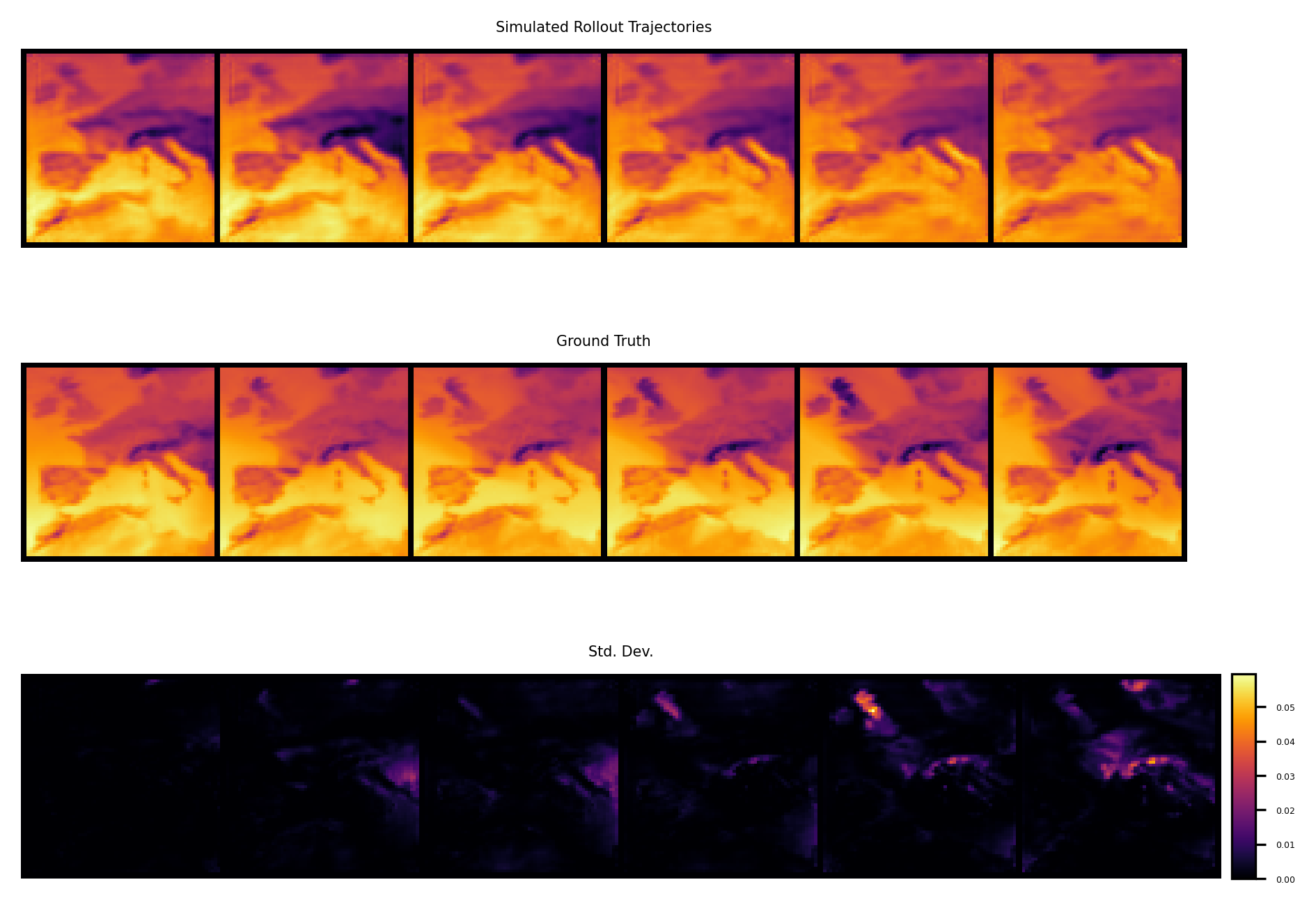}
  \text{$\qquad$ $\qquad$\tiny \textit{Temperature}}
 \endminipage 
 \minipage{0.4\columnwidth}
   \includegraphics[width=\columnwidth]{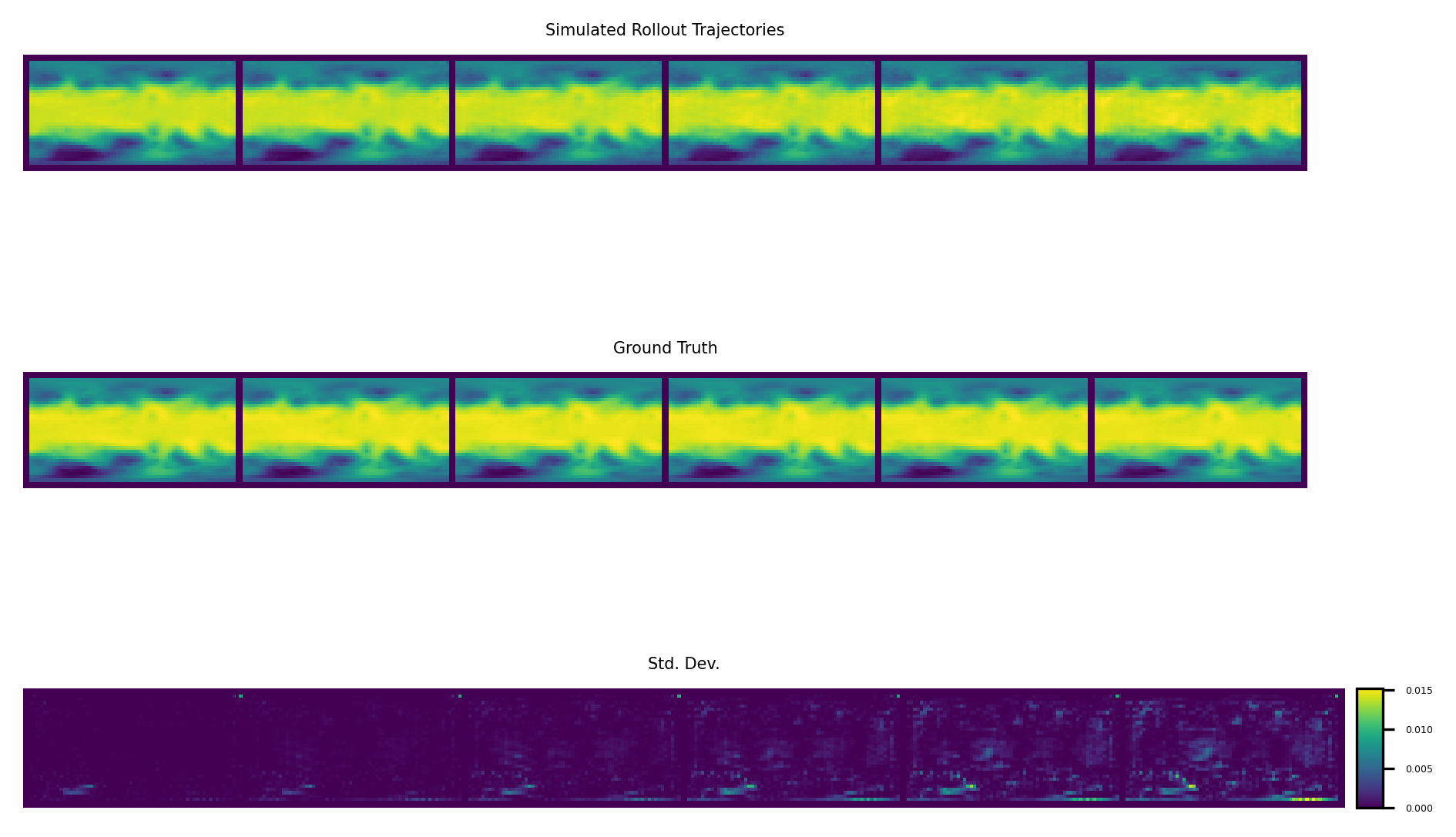}
   \text{$\qquad$$\qquad$ \tiny \textit{Geopotential}}
 \endminipage 
 \caption{\textit{Visualization of rollout trajectories starting from the same initial conditions on the ERA5 temperature and geopotential dataset from the \textbf{3DUNet} model. The last row shows the squared absolute error. For rollouts from other methods, see appendix.}}
 \label{fig:rolloutsunet}
 \end{figure}

\section{Appendix: Architecture} \label{appdx:architecture}
\subsection{Spatio-Temporal Conditional Normalizing Flow}



\begin{table}[h]
\caption{\textit{Architecture details for a single coupling layer in the spatio-temporal flow architecture. The variable $c_{\mathrm{out}}$ denotes the number of output channels. All convolutional layers are followed by a ReLU activation.}}
    \label{tab:coupling_architecture_sr}
    \centering
        \resizebox{0.5\columnwidth}{!}{%
    \begin{tabular}{c | c | c}
        \textbf{Layer} & \textbf{Hidden Channels } & \textbf{Kernel size} \\
        \midrule
        Upsample Conv3d & 1 & $1 \times 1 \times 1$ \\
        Conv3d & 512 &$3\times 3$ \\
        Conv3d  & 512 & $3 \times 3$ \\
        Conv3d & $c_{\mathrm{out}}$ & $1 \times 1$ \\
        \bottomrule
    \end{tabular}}
\end{table}

\subsection{Gated Convolutional LSTM}

\begin{table}[h]
\caption{\textit{Architecture details for the full gated convolutional neural network, as described in Figure \ref{fig:gated-conv-net}. We chose n=6 layers for all experiments.}}
    \label{tab:gated-conv-net-hparams}
    \centering
    \resizebox{0.5\columnwidth}{!}{%
    \begin{tabular}{c | c | c | c}
      \textbf{Layer} &\textbf{Hidden Channels} & \textbf{Kernel size} & \textbf{Padding}\\ 
       \midrule
       First Conv3d & 128 & 3x3  & 1 \\
       Last Conv3d & 256 & 3x3  & 1 \\
        \bottomrule
    \end{tabular}}
\end{table}

\begin{table}[h]
\caption{\textit{Architecture details for the gated convolutional layer, as described in Figure \ref{fig:gated-conv-block}. The first convolutional layer is followed by a concatenated ReLU activation.}}
    \label{tab:feature_architecture}
    \centering
    \resizebox{0.5\columnwidth}{!}{%
    \begin{tabular}{c | c | c | c}
      \textbf{Layer} &\textbf{Hidden Channels} & \textbf{Kernel size} & \textbf{Padding}\\ 
       \midrule
       Conv3d & 128 & 3x3  & 1 \\
       Conv3d & 256 & 3x3  & 1 \\
        \bottomrule
    \end{tabular}}
\end{table}

\begin{figure}[]
\centering
\includesvg[width=0.3\columnwidth]{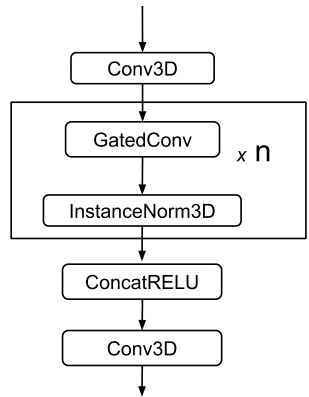}
\caption{\textit{Schematic of full gated convolutional neural network architecture with \textit{n} layers.}}\label{fig:gated-conv-net}
\end{figure}

\begin{figure}[]
\centering
\includesvg[width=0.3\columnwidth]{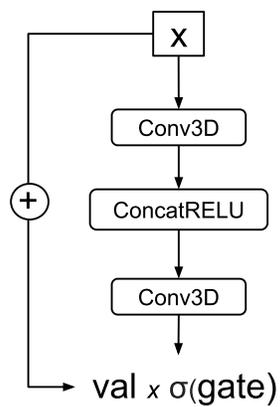}
\caption{\textit{Schematic of gated convolutional layer. 'val' stands for output value of the layer.}}\label{fig:gated-conv-block}
\end{figure}


\end{document}